\pdfoutput=1
\documentclass[11pt,a4paper]{article}
\PassOptionsToPackage{breaklinks}{hyperref}
\usepackage[hyperref]{eacl2021}

\usepackage{times}
\usepackage{latexsym}
\usepackage[T1]{fontenc}

\usepackage{microtype}

\usepackage{listings}
\usepackage{amsmath}
\usepackage{booktabs}

\usepackage{algorithm}
\usepackage{algpseudocode}

\algtext*{EndWhile}
\algtext*{EndIf}
\algtext*{EndFor}
\algtext*{EndFunction}

\usepackage{amsmath,amsfonts,amssymb}
\usepackage[htt]{hyphenat}
\usepackage{graphicx}

\aclfinalcopy %

\title{Named Entity Recognition and Linking\\Augmented with Large-Scale Structured Data}

\author{Pawe{\l} Rychlikowski \\
  University of Wroc{\l}aw \\
  \texttt{prych@cs.uni.wroc.pl} \\
  \\\And
  Bart{\l}omiej Najdecki \\
  University of Wroc{\l}aw \\
  \texttt{bnajdecki@gmail.com} \\
  \\\AND
  Adrian {\L}a{\'n}cucki \\
  NVIDIA Corporation\\
  \texttt{alancucki@nvidia.com} \\
  \\\And
  Adam Kaczmarek \\
  VoiceLab AI \\
  \texttt{adam.kaczmarek@voicelab.ai} \\
}

\date{}

\begin{document}
\maketitle
\begin{abstract}
In this paper we describe our submissions
to the 2nd and 3rd
SlavNER Shared Tasks held at BSNLP 2019 and BSNLP 2021, respectively.
The tasks focused on the analysis of Named Entities in multilingual Web documents in Slavic languages
with rich inflection. %
Our solution takes advantage of large collections
of both unstructured and structured documents. The former serve as data for unsupervised training
of language models and embeddings of lexical units.
The latter refers to Wikipedia and its structured counterpart - Wikidata,
our source of lemmatization rules, and real-world entities.
With the aid of those resources, our system could recognize, normalize
and link entities, while being trained with only small amounts of labeled data.
\end{abstract}

\section{Introduction}
Intelligent analysis of texts written in natural languages,
despite the advancements made with deep neural networks,
is still regarded as challenging.
The \emph{lingua franca} of science is English, and new methods are typically evaluated firstly on English data,
and often on other Germanic or Romance languages.
This puts a certain bias on the development and design of modern NLP methods,
which are not always transferable, and the metrics comparable, across languages and language families.

Due to the complexity and inherent vagueness of intelligent language processing,
is has been naturally split into simple tasks, one of which is named entity recognition (NER),
concerned in this paper.
The output of a NER system is traditionally a set labelled phrases recognized in a given text.
In order to process a document, one has to not only find and label the entities, but also link appropriately
subsequent occurrences of the same entity. The task becomes harder, if the linking can be made across languages,
when the entities are globally present.

We describe our submission to the
\emph{3rd Multilingual Named Entity Challenge in Slavic languages},
held at the 8th Workshop on Balto-Slavic Natural Language Processing (BSNLP)
in conjunction with the EACL 2021 conference.
The system was similar to the one submitted to the
\textit{2nd Multilingual Named Entity Challenges in Slavic languages} \cite{piskorski_second_2019}
held at 7th BSNLP Workshop in conjunction with ACL 2019 conference, and we discuss the differences between both systems.

The aim of those shared tasks was to recognize, normalize, and ultimately link -
on a document, language and cross-language level - all named entities in collections
of documents concerning the same topic, e.g., the 2020 US presidential election.
Named entities have been split into five categories: PER (person), LOC (location), ORG (organization), PRO (product), and EVT (event).
The 2019 edition featured four Slavic languages (Czech, Russian, Bulgarian, Polish), and the 2021 edition featured six languages (the previous four plus Ukrainian and Sloven).

In our solution we have combined models trained unsupervised on large datasets,
and fine-tuned on small ones in a supervised way,
with simple, white-box algorithms that perform later stages of processing in a stable and predictable manner.
In addition, we have taken advantage of similarities between certain languages in order to augment the data
and further improve the results.

\section{Our Approach}
Our system chains three modules for named entity recognition,
lemmatization, and linking,
which correspond to the objectives of the BSNLP Shared Task.
We describe them in detail in the following sections.
Our submissions for the 2019 and the 2021 shared tasks were similar,
and differed only in the first element of the chain - the entity recognition method.

\subsection{Recognition}

\subsubsection{Additional Training Data}
Because the training datasets were small, we looked for other labeled datasets.
There is no common standard of labelling NER datasets,
and those extra datasets had to be remapped into the label set of the shared task.
However, their addition did improve the recognition scores,
and we describe them in the following paragraphs.

\paragraph{PL}
We used 1343 documents from KPWr with Named Entity annotations pre-processed with \verb"liner2-convert" \cite{Liner2} tool, flattening and mapping original categories as shown in Table~\ref{tab:kpwrMapping}. 

\begin{table}[tb]
    \caption{Class label mapping to the shared task label set in additional training datasets: KPWr \cite{KPWr}, CNEC \cite{CNEC}, and FactRuEval \cite{starostin_factrueval_2016}}
    \label{tab:kpwrMapping}
    \resizebox{\columnwidth}{!}{%
    \begin{tabular}{ p{0.1\columnwidth} p{0.1\columnwidth} p{0.9\columnwidth}}
      \toprule
KPWr& PER & \texttt{nam\_adj\_person}, \texttt{nam\_liv\_*}\\
    & LOC & \texttt{nam\_adj\_city}, \texttt{nam\_oth\_address\_street}, \texttt{nam\_fac\_*}, \texttt{nam\_loc\_*}\\
    & EVT & \texttt{nam\_eve\_*}\\
    & PRO & \texttt{nam\_oth\_tech}, \texttt{nam\_pro\_*}, \texttt{nam\_oth\_license}, \texttt{nam\_oth\_stock\_index}\\
    & ORG & \texttt{nam\_org\_*}\\
    \midrule
CNEC& PER & \texttt{p} (personal names)\\
    & LOC & \texttt{g} (geographical names)\\
    & EVT & \texttt{ia} (conferences), \texttt{tc} (centuries), \texttt{tf} (feasts), \texttt{tp} (epochs)\\
    & PRO & \texttt{cs} (article titless), \texttt{mn} (periodicals), \texttt{oa} (cultural artifacts), \texttt{op} (products), \texttt{or} (directives) \\
    & ORG & \texttt{ic} (cultural/edu/science institutions), \texttt{if} (companies), \texttt{io} (govt. inst.), \texttt{mt} (tv stations)\\
    \midrule
FactRu& PER & \texttt{name}, \texttt{surname}, \texttt{nickname}, \texttt{patronymic}\\
      & LOC & \texttt{geo\_adj}, \texttt{loc\_descr}, \texttt{loc\_name} \\
    & PRO & \texttt{job}, \texttt{prj\_name}, \texttt{prj\_desc} \\
    & ORG &  \texttt{facility\_descr}, \texttt{org\_descr} \\
      \bottomrule
    \end{tabular}%
}
\end{table}

\paragraph{RU, BG, UK} For languages with Cyrillic script we used FactRuEval2016 \cite{starostin_factrueval_2016} corpus consisting of 255 documents with 11754 annotated spans. Interestingly, the addition of this dataset improved scores for BG and UK despite the language mismatch.

\paragraph{CS, SL}
For Czech and Slovene we used Czech Named Entity Corpus \cite{CNEC} containing 8993 sentences with manually annotated 35220 named entities, classified according to a two-level hierarchy.

\subsubsection{Flair-based Recognition System}

Recognition in our 2019 submission was realized with Flair \cite{akbik-etal-2018-contextual},
a model made of the embedding layer and a bi-directional LSTM with a Conditional Random Field output (BiLSTM-CRF).
The embedding layer aggregated pre-trained word representations of varying granularity and origin
(word embeddings, subword embeddings \cite{heinzerling_bpemb_2018}, contextual \emph{forward} and \emph{backward}
character embeddings inherent to Flair).

Because of the data scarcity, we adopted the philosophy of making our systems
,,neural gazetteers''. To this end, we tried to collect as much various embeddings as possible.
This line of reasoning applied especially to word-level embeddings.
Ideally we wanted our systems to have, for every language, embeddings trained on Wikipedia,
Common Crawl\footnote{\url{http://commoncrawl.org}}, and a collection of news articles.

We found it beneficial to mix word pieces and character embeddings between languages.
For instance, our model for Russian used Bulgarian embeddings.This is especially useful
when the model of specific granularity in the target language is unavailable.
Lastly, we also found it beneficial to mix training data for seemingly related languages,
and improved the scores by adding our additional FactRuEval data to the Bulgarian training dataset.

Our recurrent recognition model underperformed in comparison to the top 2019 contestants,
notably those based on BERT \cite{arkhipov_tuning_2019,devlin-etal-2019-bert}.
We present an excerpt from the 2019 recognition results in Section~\ref{sec:resultsRecon}.

\subsubsection{FLERT-based Recognition System}
For our submission to the 2021 BSNLP Shared Task
we have used FLERT \cite{schweter_flert_2020}, a state-of-the-art architecture for named entity recognition.
It is a BERT-style transformer approach, in which a XLM-RoBERTa model \cite{conneau_unsupervised_2019},
initially trained on a 100-language Common Crawl corpus \cite{wenzek_ccnet_2020},
is fine-tuned on a small, language-specific corpus.
This model departs from training an output CRF.
We found that FLERT models train fast, and outperform our previously used Flair models by a significant margin.

\subsection{Lemmatization}
In the process of lemmatization of compound phrases, some words are converted into their lemmas,
and some words remain unchanged. Occasionally some words are changed into other forms, e.g., 
adjectives might be transformed to nominatives with an appropriate gender.
In the low-data regime of the shared task, we have opted for a simple
rule-based system and data augmentation.

We pose the lemmatization task as splitting a word $w$ into two concatenated parts $w=w_1 w_2$,
and computing the lemma as $w_1 v_2$, where $(w_2, v_2) \in R_{\mathit{lem}}$,
and $R_{\mathit{lem}}$ is a small set of single-word lemmatization rules. 

We use two main additional sources of information:

\paragraph{Wikipedia} We take advantage of numerous links between articles,
from which we extract pairs \verb+[[text anchor|document title]]+ .
The anchors often are the inflected forms,
and document titles the lemmatized forms of the same entity.
In order to filter out spurious
we consider a pair (\emph{anchor}, \emph{title}) a correct lemmatization
if both the anchor and the text have the same number of words,
and every $i$-th word in a title is either equal to the $i$-th word in an anchor, or is its possible lemma.

Finally, we heuristically recognize a small set of of words for later use, which we call \textbf{stopper words}.
We define them as words shared between the anchor and the title,
such that all words that follow them are identical in the anchor and the title, e.g.,
in the (\emph{anchor}, \emph{title}) pair
\begin{align*}
\text{(}&\text{\emph{Bazylik\k{e} \underline{\`{s}w.} Paw\l{}a za Murami},}\\
        &\text{\emph{Bazylika \underline{\`{s}w.} Paw\l{}a za Murami}),}
\end{align*}
a stoper word is \emph{"\`{s}w."}.

\paragraph{Universal Dependencies}
(UD) \cite{ud} is a large collection of treebanks in multiple languages.
We extract morphosyntactic information (word, lemma, POS-tags and additional parameters\footnote{We take the 'international version' of these parameters})
from the words present in UD subsets for our target languages.
Using that information, we construct single-word lemmatization rules.
We say that the word $w$ is a possible lemma of $v$ if there is a one word lemmatization rule transforming $v$ into $w$.

\paragraph{PoliMorfologik}
For the Polish language, we additionally use PoliMorfologik \cite{wolinski_polimorf_2021},
a comprehensive morphosyntactic dictionary,
which allows us to extract a large collection of lemmatization rules.

\subsubsection{Lemmatization Schemas}
Lemmatization of every phrase gives rise to a lemmatization schema.
It works as follows: for every word we take its suffix (the longest suffix which occurs in the list of 2000 most popular suffices),
in that way we obtain the left-hand side of the rule. The right-hand side describes, how this suffices should be transformed.
For instance for the pair
\begin{align*}
  \text{(\emph{V\'{a}clavem Havlem, V\'{a}clav Havel})}
\end{align*}
we obtain a rule
\begin{align*}
  \text{(\emph{-vem, -vlem})} \quad\longrightarrow\quad \text{(\emph{-v, -vel}).}
\end{align*}

Our lemmatization algorithm takes a phrase (named entity found in the first stage) and returns its lemma.
It follows that we do not consider every information from the words surrounding the phrase/context.
Afterwards, we try to apply the following heuristics in a given order:

\begin{enumerate}
    \item Try to find the (rightmost) stopper word. If there is one, then leave unchanged suffix of the phrase after the stopper (including the stopper itself), find the lemma for the prefix.
    \item Try to apply rule based agreement phrase lemmatization (only for Polish)
    \item Try to find the lemmatization schema suitable for the phrase. If there are more than one such rule, use the one which gives ,more natural lemmatization' (which prefers common words and words occurring in lemmas)
    \item Replace every word with its most popular lemma (in the training data, and in Wikipedia), if the word doesn't occur leave it unchanged
\end{enumerate}

\subsection{Entity Linking}
A recognized entity, associated with a category and a normalized lemma,
has to be linked with other occurrences of this entity (in this document, in other documents,
and ultimately across the documents in all languages).
The task is difficult due to the subtle differences between seemingly identical entities.
Consider \emph{Donald Trump} entity: its one occurrence could be linked with \emph{the 45th president of the United States},
or \emph{Donald Trump Jr}, depending on the role in the text, but not with both at the same time.

We divide the task into two phases: 1) initial assignment of identifiers, and 2) refinement of identifiers.
Our linking algorithm relies on three kinds of matches: exact matches of entity names, partial matches,
and fuzzy matches with word embeddings.
In order to ground the recognized entities regardless of the language, as well as extend our inventory of entities
and their possible names, we use Wikidata\footnote{\url{http://www.wikidata.org}} as a catalogue of entities.

\subsubsection{Wikidata}
Wikidata is a structured database of entities extracted from Wikipedia.
Every entity has a unique identifier, e.g. \texttt{Q123456},
a list of labels and languages for each label,
a description and subclasses/instances of properties,
and relationships to other Wikidata entities (\emph{instance of}, \emph{part of}, etc.),
which form a graph.

Thanks to the hierarchy of the relations,
we have selected a handful of top-level Wikidata entities (Table~\ref{tab:wikidata}),
and collected all their descendants into sets of \emph{wikidata\_entities}.
These are further weighted by their Term Frequency in Wikidata,
so we could resolve collisions in favor of the most popular entities.

\begin{table}
\caption{Relations between Wikidata categories and named entity categories}
\label{tab:wikidata}
\resizebox{\columnwidth}{!}{%
\begin{tabular}{ p{0.1\columnwidth} p{1.0\columnwidth}}
  \toprule
  Label & Top-level Wikidata Entities\\
  \midrule
  PER & human (Q5), nationality (Q231002), ethnic group (Q41710)\\
  LOC & locality (Q3257686), location (Q2221906), spatial entity (Q58416391), geologic province (Q214045)\\
  EVT & event (Q1656682), social phenomenon (Q602884), occurrence (Q1190554)\\
  PRO & type of manufactured good (Q22811462), tangible good (Q1485500), broadcasting program (Q11578774), intellectual work (Q15621286), television station (Q1616075)\\
  ORG & organization (Q43229), trade agreement (Q252550), company (Q783794)\\
  \bottomrule
\end{tabular}%
}
\end{table}

\subsubsection{Initial Assignment of Identifiers}
In a typical, coherent paragraph, the narrative develops with every new sentence.
Upon introduction, the entities are named carefully (e.g., with a full name, expanded acronym),
to be shortened later, when it is clear from the context what they refer to.
For this reason we designed a stateful algorithm, that processes and refines a local list of \emph{doc\_entities}
caught in the document.

Algorithm 1 outlines the linking procedure.
Assignment of identifiers is performed separately for every document
with the \textsc{add\_and\_link} function. It processes a lemmatized set of entities recognized
earlier modules of our system.
Two kinds of entity dictionaries: \emph{doc\_entities}, which is local to a function,
and a global \emph{wikidata\_entities}, which we prepare earlier using Wikidata. 
Those dictionaries map the textual mentions to identifiers from Wikidata and the target language,
e.g., \emph{Donald Trump} maps to \texttt{[(Q22686, en), (Q22686, pl), (Q22686, cs), (Q3713655, cs)]}
(the last identifier refers to \emph{Donald Trump Jr}).

We process document entities starting from the longest ones,
and for each select the best entity id with the \textsc{best\_id} function.
It firstly prefers the matching entries from the \emph{doc\_entities} dictionary,
and secondly the most popular Wikidata entries (by Term Frequency) from \emph{wikidata\_entities}.
For instance, with the local \emph{doc\_entities} dictionary,
after processing \emph{Donald Trump}, a subsequent shorter mention \emph{Trump} should be linked with it.

The function \textsc{aliases} handles only PRO and ORG labels, and returns a list of all short forms
and acronyms specific to those labels, present in Wikidata, 
e.g., \emph{Sony Ericsson} is aliased as \emph{SE}.

\begin{algorithm*}[tb]
\begin{algorithmic}
\label{alg:add_identifiers}
\Function{add\_and\_link}{$\mit{ners}$}
  \State  $\mit{doc\_entities}, \mit{linked} \gets \{\}, \{\} $ 
  \For{($\mit{phrase}$, $\mit{lemma}$, $\mit{type}$) \textbf{in} \textsc{sorted}($\mit{ners}$)}
  \Comment{Descending by the \# of words in a phrase}
      \State{ $P_1 \gets \textsc{get\_identifiers}(\mit{phrase}, \mit{doc\_entities})$}
      \State{ $P_2 \gets \textsc{get\_identifiers}(\mit{lemma}, \mit{doc\_entities})$}
      \State{ $\mit{id} \gets \textsc{best\_id}(\mit{lemma}, P_1 + P_2 + [\mit{lemma} + \mit{type}]$)}
      \State $\mit{linked}[(\mit{phrase}, \mit{lemma}, \mit{type})] \gets \mit{id}$
      \State $\mit{doc\_entities} \gets \mit{doc\_entities} \cup \textsc{aliases}(\mit{lemma}, \mit{id})$

  \EndFor
  \State \textbf{return} $\mit{linked}$
\EndFunction
\\
\Function{get\_identifiers}{$\mit{phrase}$, $\mit{doc\_entities}$}
    \State{$\mit{res} \gets []$}
    \For{($\mit{doc\_phrase}$, $\mit{id}$) \textbf{in} $\mit{doc\_entities}$}  \Comment{IDs matching in the document}
        \If{\textsc{same\_entity}($\mit{doc\_phrase}, \mit{phrase}$) }
            \State{\textsc{append}($\mit{res}, \mit{id}$)}
        \EndIf
    \EndFor
    \If{$\mit{phrase} \in \mit{wikidata\_entities}$} \Comment{The most common ID for a phrase}
        \State{\textsc{append}($\mathit{res}$, $\mathit{wikidata\_entities[lemma]}$)}
    \EndIf
    \State \Return $\mit{res}$
\EndFunction
\end{algorithmic}
\caption{Basic routines of the linking algorithm}
\end{algorithm*}

\subsubsection{Refinement of Identifiers}
The refinement stage uses dense embeddings of phrases in order to uncover
high similarities between them, that might have been otherwise missed. 
We use FastText \cite{bojanowski_enriching_2017}, which is suited
for morphologically rich Slavic languages, since the representations are built from generic subword units.

The refinement is carried out in two phases.
In the first one,
all phrases with the same identifier are grouped together.
In the second one,
two groups are merged into one if there exist two mentions (one per each group)
with sufficiently similar embeddings
measured by their dot product.
Phrases are embedded as sums of embeddings of their words.
When we merge two groups, we assign to them the identifier with a higher Wikidata term frequency.
We refine identifiers only on the single language level.

\section{Evaluation}
We present experiments carried out on different levels of the entity recognition pipeline.
The data used in those experiments comes from the BSNLP 2019 Shared Task test set
(\emph{Nord Stream} and \emph{Ryanair} subsets).
Our algorithms are tested in the submitted form and have not been further adapted to those datasets.

\subsection{The 2019 Shared Task}
\label{sec:resultsRecon}

\paragraph{Recognition}
Table~\ref{tab:recon2019} summarizes \emph{strict} recognition results on the test data.

\begin{table}[h]
  \caption{2019 BSNLP Shared Task selected results (\emph{strict} recognition evaluation, test set, F1 metric).
           For every submitter, the best solution is shown with respect to the average performance on all languages.}
  \resizebox{\columnwidth}{!}{%
  \begin{tabular}{llccccc}
  \toprule
   Model      & Testset     & BG   & CS & PL & RU & All \\
   \midrule
 RIS-slav\_lemma   & NordS & 0.84 & 0.89 & 0.89 & 0.78 & 0.85\\
 CogComp-7         & NordS & 0.84 & 0.89 & 0.86 & 0.72 & 0.83\\
 IIUWR.PL-5        & NordS & 0.71 & 0.83 & 0.86 & 0.65 & 0.78\\
 TLR               & NordS & 0.73 & 0.74 & 0.72 & 0.60 & 0.70\\
 Cog\_Tech\_Cent-4 & NordS &   -  &   -  &   -  & 0.69 & 0.69\\
 Sberiboba         & NordS & 0.63 & 0.71 & 0.68 & 0.60 & 0.66\\
 JRC-TMA-CC-4      & NordS & 0.67 & 0.50 & 0.42 & 0.52 & 0.52\\
 NLP\_Cube        & NordS & 0.14 & 0.16 & 0.09 & 0.11 & 0.12\\
 \midrule
 CogComp-6         & Ryanair & 0.88 & 0.94 & 0.91 & 0.94 & 0.92\\
 RIS-slav\_lemma   & Ryanair & 0.86 & 0.94 & 0.92 & 0.91 & 0.91\\
 Cog\_Tech\_Cent-4 & Ryanair &   -  &   -  &   -  & 0.91 & 0.91\\
 IIUWR.PL-4        & Ryanair & 0.76 & 0.87 & 0.84 & 0.79 & 0.82\\
 TLR               & Ryanair & 0.76 & 0.83 & 0.82 & 0.83 & 0.82\\
 Sberiboba         & Ryanair & 0.65 & 0.84 & 0.81 & 0.72 & 0.77\\
 JRC-TMA-CC-1      & Ryanair & 0.64 & 0.55 & 0.52 & 0.79 & 0.64\\
 NLP\_Cube         & Ryanair & 0.15 & 0.13 & 0.19 & 0.18 & 0.16\\
  \bottomrule
  \end{tabular}%
  }
  \label{tab:recon2019}
\end{table}

\paragraph{Lemmatization}
We analyzed the influence of various part of lemmatization on the performance of our method.
The results are shown in Table~\ref{tab:lemmatization}.
Our baseline is the identity function, in which we assume a phrase being its own lemma. 

\begin{table}[h]
\caption{Accuracy of our rule-based lemmatization algorithm on the 2019 BSNLP Shared Task training data.
Abbreviations: \textbf{p} -- phrase lemmatization rules,
\textbf{w} -- separate lemmatization of words,
\textbf{W} -- additional Wikipedia data,
\textbf{a} -- handwritten agreement rules (Polish only),
\textbf{s} -- uses \emph{stoper words}.}
\resizebox{\columnwidth}{!}{%
\begin{tabular}{l c c c c c}
\toprule
  Method & BG & CS & PL & RU & Avg \\
\midrule
Baseline    & 89.02 & 59.23 & 54.51 & 54.79 & 63.00       \\
+a          & 89.02 & 59.23 & 58.53 & 54.79 & 64.12       \\
+w          & 89.91 & 64.39 & 74.17 & 57.23 & 70.41       \\
+p          & 89.18 & 67.47 & 79.12 & 57.53 & 72.34       \\
+wW         & 92.73 & 71.29 & 81.27 & 86.71 & 82.62       \\
+pW         & 88.53 & 81.78 & 80.77 & 89.16 & 84.97       \\
+paswW      & 91.60 & 81.69 & 82.42 & 89.28 & 86.14       \\
+pasW       & 92.33 & 81.86 & 83.57 & 89.99 & {\bf 86.83} \\
\bottomrule
\end{tabular}%
}
\label{tab:lemmatization}
\end{table}
One should be aware that due to the small amount of test data, the results should be treated as approximate.
Some differences can be caused by bad lemmatization of one phrase
(especially if the phrase occurs many times in test data).
It seems that all implemented heuristic are reasonable and improve over the baseline.
Moreover, it is easy to see that links from Wikipedia are useful source of information in this task.

\paragraph{Entity Linking}
Table~\ref{tab:linkingOracle} shows the result of linking. Even though our recognizer did not hold up to the competition,
the linking algorithm was able to close the gap in F1 score.
In order to test the algorithm in ablation, we include linking results on ground truth lemmatized data (Lemma Oracle).

\begin{table}[htb]
\caption{2019 BSNLP Shared Task results (cross-language linking, test data).
For every team we present their highest scoring submission wrt. the F1 metric.
(*) The oracle model (first row for every dataset) is our entity linking
algorithm run on the ground truth lemmatized data after the competition.}
\resizebox{\columnwidth}{!}{%
\begin{tabular}{llccc}
\toprule
 Model      & Testset     & F1   & Prec. & Rec. \\
 \midrule
Ours + Lemma Oracle & Ryanair & 0.76$^{*}$ & 0.83$^{*}$ & 0.70$^{*}$\\
Ours (IIUWR.PL-5) & Ryanair & 0.49 & 0.80 & 0.35\\
JRC-TMA-CC-2      & Ryanair & 0.27 & 0.67 & 0.17\\
CogComp-3         & Ryanair & 0.13 & 0.07 & 0.73\\
RIS-merge         & Ryanair & 0.10 & 0.06 & 0.70\\
Sberiboba         & Ryanair & 0.10 & 0.06 & 0.30\\
NLP\_Cube          & Ryanair & 0.00 & 0.67 & 0.00\\
\midrule
 Ours + Lemma Oracle & NordS & 0.59$^{*}$ & 0.74$^{*}$ & 0.50$^{*}$ \\
 Ours (IIUWR.PL-5)  & NordS & 0.42 & 0.73 & 0.29 \\
 JRC-TMA-CC-2       & NordS & 0.31 & 0.69 & 0.20 \\
 RIS-merge\_lemma   & NordS & 0.11 & 0.06 & 0.72 \\
 CogComp-3          & NordS & 0.11 & 0.06 & 0.68 \\
 Sberiboba          & NordS & 0.06 & 0.03 & 0.36 \\
 NLP\_Cube          & NordS & 0.00 & 0.46 & 0.00 \\
\bottomrule
\end{tabular}%
}
\label{tab:linkingOracle}
\end{table}

\subsection{The 2021 Shared Task}
We present the results of our FLERT-based submission,
which are partial results of the entire shared task available at the time of writing.

One of the sets of articles in the training data is devoted to COVID-19. This situation is unusual: the phrase very often used in test data, does not appear at all in the training data (also in the data used to pre-train language model).

We have verified that our NER models struggle with assigning consistent labels to the phrase \emph{COVID-19},
which is common in the test data.
An additional difficulty is the ambiguity of this phrase, which may refer to a disease
and possibly remain unclassified as a named entity, or a pandemic and be classified as EVT.
We decided to do a simple post-processing which assigns EVT to all COVID-19 phrases recognized by the NER module. 

We think that this situation is so unusual that in a real system, used in the industry,
it would be handled using a special ad-hoc rule.
Moreover, we wanted to know, what are the result of this fixed assignment, and submitted two versions of our solutions.

\begin{table}[htb]
  \caption{2021 BSNLP Shared Task selected results (test set, F1 metric):
           \emph{strict} recognition, normalization, language-level linking (coreference).
           NC refers to the submission without fixed labelling of COVID-19 occurrences as EVT.}
  \resizebox{\columnwidth}{!}{%
  \begin{tabular}{llccccccc}
  \toprule
Task &   Testset     &   CS &   RU &   BG &   UK &   SL &   PL &  All\\
   \midrule
Recon.   &  US Elect.   & 0.87 & 0.70 & 0.82 & 0.79 & 0.88 & 0.86 & 0.79\\
         &  COVID-19    & 0.80 & 0.57 & 0.72 & 0.75 & 0.77 & 0.81 & 0.73\\
(NC)     &  US Elect.   & 0.87 & 0.70 & 0.82 & 0.79 & 0.88 & 0.85 & 0.78\\
         &  COVID-19    & 0.80 & 0.55 & 0.68 & 0.72 & 0.75 & 0.78 & 0.71\\
   \midrule
Norm.    &   US Elect.   & 0.52 & 0.26 & 0.51 & 0.26 & 0.62 & 0.62 & 0.43\\
         &   COVID-19    & 0.45 & 0.27 & 0.33 & 0.51 & 0.53 & 0.57 & 0.45\\
   \midrule
Link.    &  US Elect.   & 0.66 & 0.39 & 0.69 & 0.52 & 0.66 & 0.70 & 0.56\\
         &  COVID-19    & 0.66 & 0.39 & 0.68 & 0.61 & 0.66 & 0.73 & 0.62\\
(NC)    &  US Elect.   & 0.66 & 0.39 & 0.68 & 0.52 & 0.66 & 0.70 & 0.55\\
         &  COVID-19    & 0.66 & 0.39 & 0.67 & 0.61 & 0.66 & 0.72 & 0.62\\
  \bottomrule
  \end{tabular}%
  }
  \label{tab:recon2021}
\end{table}

\section{Conclusion}
This paper describes our submissions to the 2019 and 2021 BSNLP Shared Tasks
on named entity recognition on Slavic languages.
Even though the training data was scarce, we have used large-scale datasets:
corpora of unstructured text in the unsupervised training phase of training of the recognition model,
and structured Wikipedia and Wikidata knowledge bases in order to extract rules and entities
for lemmatization and linking phases.
The linking algorithm is a strong point of our submission.
In the 2019 task it allowed to close the performance gap between our solution and competitors,
introduced by a weak initial recognition model.
The results suggest that, perhaps, there is still a white spot in between supervised and unsupervised neural learning,
where the structure of the data matters more than volume, and simple rule-based system excel.

\section*{Acknowledgment}
The authors thank Polish National Science Center for funding
under the OPUS-18 2019/35/B/ST6/04379 grant.
We also would like to thank Adam Wawrzy{\'n}ski and Wojciech Janowski from VoiceLab AI for their support
during conducting experiments and model training.

\bibliographystyle{acl_natbib}
\bibliography{refs}

\end{document}